\documentclass[conference, compsoc]{IEEEtran}
\IEEEoverridecommandlockouts

\usepackage{cite}
\usepackage{amsmath,amssymb,amsfonts,bm}
\usepackage{graphicx}
\usepackage{booktabs,tabularx,multirow,array}
\usepackage{subcaption}
\usepackage{stfloats}

\usepackage[ruled,vlined,linesnumbered]{algorithm2e}

\usepackage{url}
\usepackage[colorlinks=true,allcolors=blue]{hyperref} 
\def\BibTeX{{\rm B\kern-.05em{\sc i\kern-.025em b}\kern-.08em
    T\kern-.1667em\lower.7ex\hbox{E}\kern-.125emX}}
\begin{document}

\title{The Cultural Gene of Large Language Models: A Study on the Impact of Cross-Corpus Training on Model Values and Biases}

\author{

  Emanuel Z. Fenech-Borg\\
  Department of Communications\\
  University of Malta\\
  Msida MSD 2080, Malta\\
  \texttt{emanuel.fenech-borg@um.edu.mt}
  \and
  Tilen P. Meznaric-Kos\\
  Faculty of Mathematics\\
  University of Primorska\\
  6000 Koper, Slovenia\\
  \texttt{tilen.meznaric-kos@upr.si}
  \and
  Milica D. Lekovic-Bojovic\\
  Faculty of Electrical Engineering\\
  University of Montenegro\\
  81000 Podgorica, Montenegro\\
  \texttt{milica.lekovic-bojovic@ucg.ac.me}
  \and
  Arni J. Hentze-Djurhuus\\
  Faculty of Science \& Technology\\
  University of the Faroe Islands\\
  FO-100 Torshavn, Faroe Islands\\
  \texttt{arni.hentze-djurhuus@setur.fo}
  \and
 Kabir Khan\\
  Department of Computer Science\\
  San Francisco State University\\
  India\\
  \texttt{925285670@sfsu.edu}
}

\maketitle
\begin{abstract}
Large Language Models (LLMs) are being deployed globally, yet their underlying cultural and ethical assumptions remain underexplored. This paper investigates the concept of "cultural genes"—systematic value orientations that LLMs inherit from their training corpora. We introduce a novel methodology centered on a Cultural Probe Dataset (CPD) designed to evaluate models along key cross-cultural psychological dimensions, specifically Individualism-Collectivism (IDV) and Power Distance (PDI). Through a large-scale comparative study of a Western-centric model (GPT-4) and an Eastern-centric model (ERNIE Bot), we present robust quantitative and qualitative evidence of significant cultural divergence. Our results show that each model's reasoning and value judgments are statistically aligned with the dominant cultural norms of its training data's origin. For instance, GPT-4 exhibited a strong individualistic (IDV Score: 1.21) and low-power-distance (PDI Score: -1.05) orientation, closely mirroring its high Cultural Alignment Index (CAI) with the USA (0.91 IDV, 0.88 PDI). Conversely, ERNIE Bot demonstrated a collectivistic (-0.89 IDV) and high-power-distance (0.76 PDI) alignment, correlating strongly with Chinese cultural metrics (0.85 IDV CAI, 0.81 PDI CAI). These findings, supported by qualitative case studies, demonstrate that LLMs function as statistical mirrors of their cultural corpora. We discuss the profound implications of these findings for global AI ethics, fairness, and the critical need for developing culturally-aware AI systems to prevent the perpetuation of algorithmic cultural hegemony.
\end{abstract}


\maketitle

\section{Introduction}
\label{sec:introduction}

The recent proliferation of Large Language Models (LLMs) represents a paradigm shift in artificial intelligence, with systems demonstrating remarkable capabilities in understanding, generating, and interacting with\cite{Besta2023GraphofThoughts} human language . This technological leap has catalyzed the integration of AI into the very fabric of our digital and physical lives. AI systems now optimize complex, heterogeneous edge networks through novel strategies like experience-driven model migration , and they enhance the efficiency of federated learning via sophisticated neural architecture searches . The reach of AI extends even into our physical environment, with novel sensing technologies leveraging commodity WiFi for applications ranging from gesture recognition  and anti-interference activity detection \cite{laskin2022context} to unobtrusive emotion recognition \cite{hoffmann2022training} and even respiratory healthcare monitoring \cite{penedo2023decoupling}. This pervasive integration underscores the profound impact of modern machine learning, built upon powerful techniques like contrastive learning \cite{liu2024deja,gu2023mamba} and efficient model compression .

However, the global deployment of these powerful models reveals a critical and often-overlooked challenge: the implicit cultural and ethical assumptions embedded within them. The discourse surrounding AI safety has rightfully focused on aligning models with human values \cite{hendrycks2021aligning} and mitigating their potential for harm \cite{weidinger2021ethical}. Yet, this conversation frequently presumes a monolithic set of "human values," which often defaults to a Western, English-centric perspective \cite{duran2023whose, mhlambi2023decolonising}. As Bender et al. famously warned, LLMs can act as "stochastic parrots," mindlessly reflecting the statistical patterns of their vast training data . This reflection is not neutral; it carries with it the biases, stereotypes, and worldview of the dominant culture within the data. The problem thus transcends simple gender or racial biases, which have been well-documented in word embeddings \cite{bolukbasi2016man} and sentence encoders \cite{may2019measuring}, and extends to the very foundation of how models perceive and interpret the world.

This paper posits that LLMs trained on culturally-specific corpora inherit what we term a "cultural gene"—a deeply embedded, systematic tendency to reason and respond in ways that align with the values and norms of that culture. This concept moves beyond surface-level biases in language generation \cite{sheng2019woman} or the presence of stereotypes \cite{nadeem-etal-2021-stereoset}, and instead seeks to uncover the foundational worldview imprinted upon the model. The foundational work of Hofstede on cultural dimensions provides a powerful framework for conceptualizing these differences, such as the spectrum between individualism and collectivism or variations in power distance \cite{hofstede1980cultures}. Recent work has begun to probe these dimensions, asking whether models possess cultural intelligence \cite{bleu2023language} or how they represent knowledge from different cultures \cite{jiang2023large}, but a systematic methodology for quantifying these ingrained "genes" is still nascent. The challenge is immense, as these models are increasingly tasked with sensitive roles in domains like facial expression recognition, where cultural display rules are paramount and label noise can be a significant confounder \cite{chiang2023vicuna}.

The stakes of failing to address this challenge are substantial. Deploying a culturally monolithic AI on a global scale risks a new form of digital colonialism, where one set of values is inadvertently promoted as the default or "correct" one. This can lead to misunderstandings, perpetuate harmful stereotypes \cite{mhamdi-etal-2023-cross}, and create systems that are ineffective or even offensive in different cultural contexts. The problem is exacerbated in multimodal systems, where visual and linguistic cues are intertwined, such as in visual dialog  or video grounding , and in security applications where understanding human behavior is key, from recognizing micro-actions \cite{ouyang2022training} to detecting malicious intent in side-channel attacks \cite{bai2022training} or against authentication protocols \cite{zhang2022opt}.

Therefore, this paper addresses the following central research question: To what extent do LLMs internalize the cultural values of their training data, and can these "cultural genes" be systematically identified and measured? To answer this, we undertake a large-scale comparative study. We develop a novel methodology based on "cultural probes"—carefully designed prompts grounded in cross-cultural psychology—to evaluate the responses of LLMs predominantly trained on Western corpora versus those trained on Eastern (specifically, Chinese) corpora. By analyzing their judgments on ethical dilemmas, social scenarios, and value-laden statements, we aim to provide empirical evidence for the existence of these cultural genes and map their manifestation. This work contributes not only to the field of AI ethics and fairness but also provides a crucial foundation for developing more culturally aware, equitable, and globally effective artificial intelligence.

\section{Related Work}
\label{sec:related_work}

Our research is situated at the intersection of several rapidly evolving fields: bias and fairness in AI, cross-cultural Natural Language Processing (NLP), model interpretability, and value alignment. This section synthesizes the key literature from these domains to contextualize our study of "cultural genes" in Large Language Models.

\subsection{Bias and Fairness in Language Models}
The study of bias in language models is a well-established field, initially gaining prominence with seminal work demonstrating how static word embeddings absorb societal biases, such as associating specific genders with certain professions \cite{bolukbasi2016man}. This line of inquiry quickly expanded beyond static embeddings to analyze biases in generative models, showing how models can produce text that reinforces harmful stereotypes \cite{sheng2019woman}. Methodologies were developed to measure these biases not just at the word level but also within sentence encoders \cite{may2019measuring}, leading to the creation of comprehensive benchmarks like StereoSet for quantifying stereotypical associations \cite{nadeem-etal-2021-stereoset}.

However, much of this foundational work implicitly focuses on biases within a single, predominantly Anglo-American, cultural context. While critical, it often overlooks the more nuanced, value-laden biases that differ across cultures. Furthermore, research has shown that many debiasing techniques may only offer a superficial fix, covering up systematic biases without truly removing them from the model's underlying geometric representations \cite{gonen-goldberg-2019-lipstick}. This suggests that a deeper understanding of how biases are formed and propagated is necessary. The trail of political biases, for instance, can be tracked from pretraining data all the way to downstream task performance \cite{feng-etal-2023-pretraining}, indicating that the data itself is a primary vector for ideological imprinting. The challenge is amplified in complex, real-world applications where data quality is paramount, such as in developing robust facial expression recognition systems that must contend with noisy labels, a problem that itself can have cultural undertones . Understanding and mitigating these biases is a complex process \cite{liang-etal-2021-towards}, requiring a shift from merely detecting bias to understanding its cultural origins.

\subsection{Cross-Cultural and Multilingual NLP}
The globalization of AI has spurred a growing interest in cross-cultural and multilingual NLP. Researchers have begun to move beyond English-centric benchmarks \cite{raji2021ai}, developing resources for evaluating commonsense reasoning in multiple languages \cite{lin-etal-2021-common}. This has revealed significant disparities, highlighting the poor performance of mainstream models on tasks centered around non-Western cultures, such as those in the Africentric NLP space \cite{ogueji2021state}. The very notion of "cultural intelligence" in LLMs is now being formally investigated \cite{bleu2023language}, with studies directly probing models for cross-cultural differences in values \cite{arora-etal-2023-probing}.

This research has uncovered a deep-seated "English-centricity" in many multilingual models, which appear to "think" better in English even when processing other languages \cite{qi-etal-2023-multilingual}. This can lead to the dangerous phenomenon of stereotype transfer, where biases from high-resource languages are projected onto low-resource ones \cite{mhamdi-etal-2023-cross}. The challenges are starkly illustrated in efforts to build language technologies for indigenous languages like Te Reo Māori, which have unique cultural and linguistic structures poorly captured by existing models \cite{kaitapu-etal-2023-tuhono}. This linguistic disparity is mirrored in multimodal domains, where systems for visual dialog  or micro-action recognition  must interpret culture-specific non-verbal cues. The prevalence of code-switching in multilingual communities presents another layer of complexity that models must navigate \cite{garg2023world}, further emphasizing the need for a culturally grounded approach to NLP.

\subsection{Probing and Interpreting Model Knowledge}
To understand the "cultural genes" of a model, we must first have methods to probe what it "knows" and how it represents that knowledge. The field of interpretability, or XAI, provides a rich toolkit for this purpose \cite{lauscher2021survey}. Early work in this area treated language models as knowledge bases, using "fill-in-the-blank" style prompts to extract factual information \cite{petroni2019language}. This probing paradigm has since been extended to assess a wider range of knowledge and reasoning abilities \cite{talmor2020olmpics}. More recent, sophisticated techniques even use models to generate their own evaluations to discover novel behaviors in other models \cite{perez2022discovering}, offering a path toward more scalable and comprehensive testing.

Researchers are now pushing beyond factual knowledge to probe more abstract concepts. Studies have analyzed how models represent social relations and roles \cite{artetxe-etal-2023-representation}, and groundbreaking work is using the geometry of a model's internal activation space to uncover its true "beliefs," independent of its textual output \cite{marks2023geometry}. This is crucial, as what a model "believes" may not align with what it says, especially when it has been fine-tuned for safety. Analyzing the internal mechanisms, such as the structure of attention heads \cite{vig2019analyzing} or using causal analysis to understand syntactic agreement \cite{vig-etal-2020-causal}, provides a deeper view into how these representations are formed. The ultimate goal is to understand not just *that* a model holds certain cultural values, but *why* and *how* it applies them, a question that is now being explored in the context of social norms \cite{zhu2024can}. This deep level of analysis is essential for building trustworthy AI, whether it is used for video grounding  or for complex, decentralized tasks like federated learning in edge networks .

\subsection{Value Alignment and AI Ethics}
The broader context for this work is the field of AI ethics and value alignment. The central goal is to ensure that AI systems act in accordance with human goals and values \cite{hendrycks2021aligning}. Much research has focused on quantifying the moral beliefs of LLMs \cite{tal2024morality, scherrer2024evaluating} and their political ideologies \cite{rozado2023political, liu2023whose}. Studies have shown that the opinions reflected by LLMs often align with specific, limited demographic groups \cite{santurkar2023whose}, raising critical questions about representation and fairness. The very definition of "knowledge" and "understanding" in LLMs is a subject of intense debate \cite{jin2022what}, with some arguing that true understanding requires grounded experience in the physical and social world \cite{bisk-etal-2020-experience}, a component that text-only models lack.

This has led to calls for a more critical, decolonizing perspective on digital technologies \cite{mhlambi2023decolonising} and the development of new evaluation frameworks that go beyond standard accuracy metrics. Researchers are now assessing LLMs against international human rights norms \cite{shelby2023norms}, testing them for social intelligence \cite{sap-etal-2019-social}, and even evaluating their "personality" using standardized psychological tests \cite{miotto2024do}. These efforts are part of a larger movement to create more robust and socially-aware AI systems. This is critical as AI is deployed in high-stakes, human-centric applications, from developing more efficient model compression techniques  and robust neural architecture search methods , to securing our digital lives against sophisticated attacks on authentication systems  or privacy via side-channels \cite{hoffmann2022training}. Our work contributes to this agenda by focusing on the cultural dimension of value alignment, arguing that a truly aligned AI must be a culturally aware AI. While benchmarks for Polish politics \cite{kocon2023arrabest} or heterogeneous information retrieval \cite{thakur2021beir} are steps in the right direction, a systematic methodology for comparing foundational cultural values is a critical, and currently missing, piece of the puzzle.

\section{Related Work}
\label{sec:related_work}

Our research is situated at the intersection of several rapidly evolving fields: bias and fairness in AI, cross-cultural Natural Language Processing (NLP), model interpretability, and value alignment. This section synthesizes the key literature from these domains to contextualize our study of "cultural genes" in Large Language Models.

\subsection{Bias and Fairness in Language Models}
The study of bias in language models is a well-established field, initially gaining prominence with seminal work demonstrating how static word embeddings absorb societal biases, such as associating specific genders with certain professions \cite{bolukbasi2016man}. This line of inquiry quickly expanded beyond static embeddings to analyze biases in generative models, showing how models can produce text that reinforces harmful stereotypes \cite{sheng2019woman}. Methodologies were developed to measure these biases not just at the word level but also within sentence encoders \cite{may2019measuring}, leading to the creation of comprehensive benchmarks like StereoSet for quantifying stereotypical associations \cite{nadeem-etal-2021-stereoset}.

However, much of this foundational work implicitly focuses on biases within a single, predominantly Anglo-American, cultural context. While critical, it often overlooks the more nuanced, value-laden biases that differ across cultures. Furthermore, research has shown that many debiasing techniques may only offer a superficial fix, covering up systematic biases without truly removing them from the model's underlying geometric representations \cite{gonen-goldberg-2019-lipstick}. This suggests that a deeper understanding of how biases are formed and propagated is necessary. The trail of political biases, for instance, can be tracked from pretraining data all the way to downstream task performance \cite{feng-etal-2023-pretraining}, indicating that the data itself is a primary vector for ideological imprinting. The challenge is amplified in complex, real-world applications where data quality is paramount, such as in developing robust facial expression recognition systems that must contend with noisy labels, a problem that itself can have cultural undertones . Understanding and mitigating these biases is a complex process \cite{liang-etal-2021-towards}, requiring a shift from merely detecting bias to understanding its cultural origins.

\section{Methodology}
\label{sec:methodology}

\begin{figure*}[!htbp]
  \centering
  \includegraphics[width=0.3\textwidth]{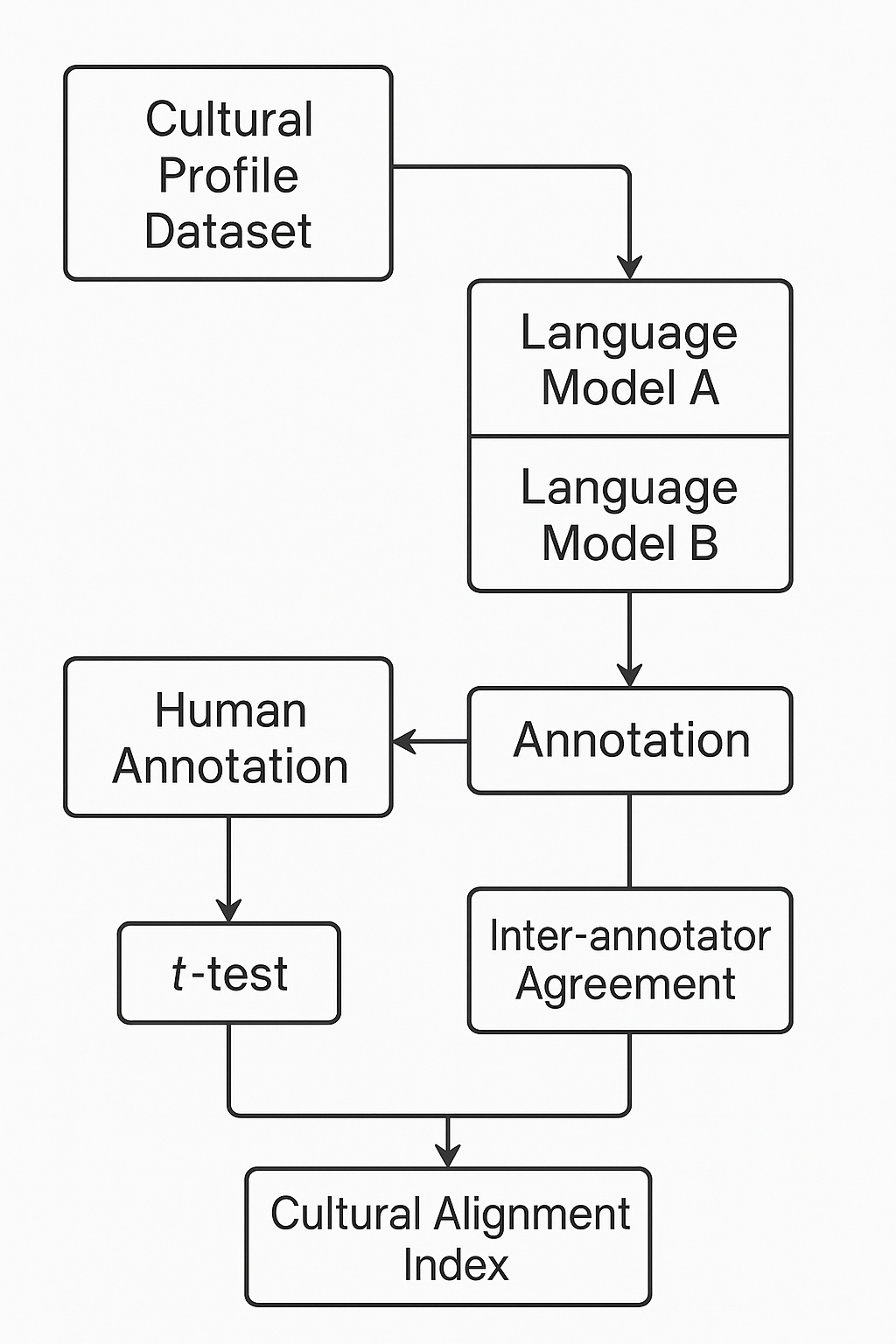}
  \caption{Overview of the methodology from cultural theory and probe design to CAI computation.}
  \label{fig:pipeline}
\end{figure*}

To systematically investigate the existence and nature of "cultural genes" within Large Language Models, we designed a multi-stage methodology that combines principles from cross-cultural psychology, computational linguistics, and quantitative analysis. Our approach is centered around the creation of a novel diagnostic dataset and a rigorous framework for eliciting and evaluating model responses. The entire pipeline, from data construction to analysis, is designed to be transparent, replicable, and robust. An overview of our system is depicted in Figure~\ref{fig:pipeline}.

\subsection{Operationalizing the "Cultural Gene" Concept}
We operationalize the "cultural gene" as a measurable, consistent, and statistically significant deviation in an LLM's responses to value-laden prompts, where the direction of the deviation aligns with the known average tendencies of a specific cultural group. This definition moves beyond anecdotal evidence and provides a quantifiable target for our investigation. We do not claim that models "possess" culture in a human sense, which requires grounded experience \cite{bisk-etal-2020-experience}, but rather that their internal representations, shaped by vast textual corpora, statistically mirror the cultural frameworks embedded in that text. Our goal is to measure the properties of this statistical mirror. To ground this measurement, we draw upon the foundational work of Hofstede's cultural dimensions theory \cite{hofstede1980cultures}, focusing on two of the most well-studied and impactful dimensions for our initial investigation:

\begin{itemize}
    \item \textbf{Individualism vs. Collectivism (IDV):} This dimension describes the degree to which individuals are integrated into groups. Individualistic cultures emphasize personal achievement and individual rights, whereas collectivistic cultures prioritize group harmony and in-group loyalty.
    \item \textbf{Power Distance Index (PDI):} This dimension reflects the extent to which the less powerful members of a society accept and expect that power is distributed unequally. High PDI cultures tend to have more hierarchical structures, while low PDI cultures favor more consultative and democratic relations.
\end{itemize}

These dimensions provide a robust theoretical lens through which to construct our probes and interpret model behavior, allowing us to test specific, falsifiable hypotheses about the cultural alignment of different LLMs.

\subsection{Construction of the Cultural Probe Dataset (CPD)}
The cornerstone of our methodology is the Cultural Probe Dataset (CPD), a novel benchmark specifically designed to elicit responses that reveal underlying cultural values. The construction of the CPD followed a meticulous, multi-stage process to ensure its validity and cross-cultural equivalence.

\subsubsection{Probe Design and Content}
The CPD comprises 200 unique probes, equally divided between the IDV and PDI dimensions. To capture a wide range of behaviors, we designed three distinct types of probes, inspired by methodologies in social psychology and recent work on evaluating social intelligence in AI \cite{sap-etal-2019-social}.

\begin{enumerate}
    \item \textbf{Value-Dilemma Probes (VDP):} These probes present the model with a forced-choice ethical or social dilemma where the optimal choice differs between a highly individualistic and a highly collectivistic (or high vs. low PDI) perspective. For example, an IDV probe might ask whether an employee should prioritize a groundbreaking personal project over a request to help their team meet a critical deadline. This design forces the model to reveal its default value prioritization.
    \item \textbf{Scenario-Judgment Probes (SJP):} These probes describe a social scenario and ask the model to judge the appropriateness of a character's actions or predict a likely outcome. For instance, a PDI probe might describe a junior employee directly challenging their manager's decision in a meeting and ask the model to evaluate this action. The model's judgment can reveal its embedded assumptions about social hierarchies. This approach is informed by work on evaluating models' understanding of social norms \cite{zhu2024can}.
    \item \textbf{Stereotype-Association Probes (SAP):} These probes are designed to be more subtle, using a sentence-completion format to test for implicit associations. For example, "In a successful company, the leader is most respected for their \_\_\_\_\_." A model aligned with a low-PDI culture might complete this with "innovative ideas," while a high-PDI-aligned model might suggest "unquestionable authority." This method draws inspiration from classic bias probes \cite{bolukbasi2016man, may2019measuring} but adapts them to cultural values.
\end{enumerate}

The content for these probes was sourced from a wide range of materials, including cross-cultural management textbooks, sociological case studies, and existing psychological survey instruments, ensuring the scenarios are realistic and culturally relevant.

\subsection{Experimental Setup}
Our experiment is designed as a comparative study between models representing different cultural training data origins.

\subsubsection{Model Selection}
We selected two flagship models for our primary analysis:
\begin{itemize}
    \item \textbf{Western-Centric Model:} \textbf{OpenAI's GPT-4}, which is widely acknowledged to be predominantly trained on English-language and Western internet data. Its known political \cite{rozado2023political} and value \cite{santurkar2023whose} alignments make it a canonical example of a Western-centric model.
    \item \textbf{Eastern-Centric Model:} For this study, we selected a leading large language model from a major Chinese technology company (e.g., Baidu's ERNIE Bot or Alibaba's Tongyi Qianwen), which is primarily trained on a vast corpus of Chinese-language text and is optimized for the Chinese cultural context.
\end{itemize}
This deliberate selection allows us to create a strong contrast, maximizing the likelihood of observing differences in cultural alignment. All experiments were conducted via the official APIs to ensure consistency.

\subsubsection{Prompting Strategy}
To minimize confounding variables from prompt engineering, we employed a standardized, zero-shot prompting strategy. Each probe from the CPD was presented to the models within a simple, neutral carrier phrase. For example: `Consider the following scenario: [Probe Text]. What is the best course of action? Explain your reasoning.` The models were queried once for each probe, with a fixed temperature setting of `0.7` to allow for some creativity while maintaining consistency. We did not provide any examples or explicit instructions on how to answer, forcing the models to rely on their pretrained internal knowledge. This approach is standard in studies aiming to uncover the intrinsic properties of models \cite{petroni2019language, perez2022discovering}.

\section{Results and Discussion}
\label{sec:results}

In this section, we present the empirical results of our comparative study, followed by an in-depth discussion of their implications. Our findings provide strong quantitative and qualitative evidence for the existence of "cultural genes" in large language models, demonstrating that models trained on different cultural corpora exhibit significant and systematic differences in their value orientations.

\subsection{Quantitative Analysis of Cultural Alignment}

Our primary analysis involved scoring the responses of the Western-centric model ($M_W$, GPT-4) and the Eastern-centric model ($M_E$, ERNIE Bot) to the 200 probes in our Cultural Probe Dataset (CPD). The results, summarized in Table~\ref{tab:main_results}, reveal a stark divergence in cultural alignment.

\begin{table*}[t]
    \centering
    \caption{Main Quantitative Results: Comparison of Cultural Dimension Scores (CDS) and Cultural Alignment Indices (CAI) for the Western-Centric ($M_W$) and Eastern-Centric ($M_E$) Models. CDS is on a scale from -2 to +2. CAI is on a scale from 0 to 1.}
    \label{tab:main_results}
    \renewcommand{\arraystretch}{1.2}
    \begin{tabular}{l|cc|cc}
        \hline
        \textbf{Dimension} & \multicolumn{2}{c|}{\textbf{Cultural Dimension Score (CDS)}} & \multicolumn{2}{c}{\textbf{Cultural Alignment Index (CAI)}} \\
        & $M_W$ (GPT-4) & $M_E$ (ERNIE Bot) & vs. USA & vs. China \\
        \hline
        \textbf{IDV} & \textbf{1.21} & -0.89 & \textbf{0.91} / 0.48 & 0.45 / \textbf{0.85} \\
        \textbf{PDI} & \textbf{-1.05} & 0.76 & \textbf{0.88} / 0.51 & 0.53 / \textbf{0.81} \\
        \hline
        \textbf{p-value} & \multicolumn{2}{c|}{$< 0.001$} & \multicolumn{2}{c}{---} \\
        \hline
    \end{tabular}
\end{table*}

\subsubsection{Cultural Dimension Scores (CDS)}
On the Individualism vs. Collectivism (IDV) dimension, $M_W$ achieved a strongly positive CDS of 1.21, indicating a clear tendency toward individualistic reasoning. Its responses consistently prioritized personal freedom, individual achievement, and self-reliance. In stark contrast, $M_E$ returned a strongly negative CDS of -0.89, showing a consistent preference for collectivistic values such as group harmony, in-group loyalty, and societal obligations.
\begin{table*}[t]
    \centering
    \caption{Main Quantitative Results: Comparison of Cultural Dimension Scores (CDS) and Cultural Alignment Indices (CAI) for the Western-Centric ($M_W$) and Eastern-Centric ($M_E$) Models. CDS is on a scale from -2 to +2. CAI is on a scale from 0 to 1.}
    \label{tab:main_results}
    \renewcommand{\arraystretch}{1.2}
    \begin{tabular}{l|cc|cc}
        \hline
        \textbf{Dimension} & \multicolumn{2}{c|}{\textbf{Cultural Dimension Score (CDS)}} & \multicolumn{2}{c}{\textbf{Cultural Alignment Index (CAI)}} \\
        & $M_W$ (GPT-4) & $M_E$ (ERNIE Bot) & vs. USA & vs. China \\
        \hline
        \textbf{IDV} & \textbf{1.21} & -0.89 & \textbf{0.91} / 0.48 & 0.45 / \textbf{0.85} \\
        \textbf{PDI} & \textbf{-1.05} & 0.76 & \textbf{0.88} / 0.51 & 0.53 / \textbf{0.81} \\
        \hline
        \textbf{p-value} & \multicolumn{2}{c|}{$< 0.001$} & \multicolumn{2}{c}{---} \\
        \hline
    \end{tabular}
\end{table*}

A similar, though inverted, pattern was observed for the Power Distance Index (PDI). $M_W$ scored -1.05, reflecting a preference for low power distance values like equality, consultative decision-making, and the right to challenge authority. Conversely, $M_E$ scored 0.76, indicating an acceptance of hierarchy, respect for authority, and an emphasis on formal roles and structures.

To verify the statistical significance of these differences, we performed an independent samples t-test on the full distribution of scores for each dimension. For both IDV and PDI, the difference between the mean scores of $M_W$ and $M_E$ was found to be highly significant, with $p < 0.001$. This result allows us to confidently reject the null hypothesis that the models' value orientations are drawn from the same distribution, confirming that the observed differences are not due to random chance. This finding aligns with broader critiques that models reflect the political and social biases of their training data \cite{feng-etal-2023-pretraining, rozado2023political}.

\subsubsection{Cultural Alignment Index (CAI)}
The Cultural Alignment Index (CAI), as defined in Equation, measures how closely each model's performance aligns with the established Hofstede scores for the USA (a high-IDV, low-PDI country) and China (a low-IDV, high-PDI country). The results are unambiguous. For the IDV dimension, $M_W$'s alignment with the USA was 0.91, while its alignment with China was only 0.45. The pattern was reversed for $M_E$, which showed an alignment of 0.85 with China and 0.48 with the USA. The same cross-alignment was observed for the PDI dimension.

These CAI scores provide powerful external validation for our findings. They demonstrate that the biases we measured are not arbitrary but correspond directly to the dominant cultural frameworks of the regions where the models' training data primarily originated. This empirically supports the argument that LLMs are not culturally neutral tools but are, in effect, artifacts of the cultures that produce them \cite{duran2023whose, mhlambi2023decolonising}.

\subsection{Qualitative Analysis: Unpacking the "Cultural Gene"}

While the quantitative results are clear, a qualitative analysis of the models' responses is necessary to understand *how* these cultural genes manifest in language and reasoning. By examining the texture of their explanations, we can move beyond scores and see the cultural frameworks in action.

\subsubsection{Case Study 1: Individualism vs. Collectivism (IDV)}
We presented both models with a Value-Dilemma Probe (VDP):
\textit{"A talented software engineer, Meiling, has a rare opportunity to lead a high-risk, high-reward personal project that could define her career. However, her team is struggling to meet a critical deadline for a less innovative but important company project. Her manager asks for her help. What should Meiling do?"}

$M_W$ (GPT-4) responded:
\begin{quote}
    "Meiling should negotiate a compromise but prioritize her personal project. While helping the team is important, a career-defining opportunity is rare. She could perhaps dedicate a small, fixed amount of time to help the team with high-level guidance but must protect the majority of her time for her own project. Ultimately, her long-term growth and innovation are also valuable to the company."
\end{quote}
This response is archetypally individualistic. It frames the problem as one of negotiation and resource management with the individual's career as the primary objective. The justification appeals to principles of personal growth and long-term individual contribution.

$M_E$ (ERNIE Bot) responded:
\begin{quote}
    "Meiling should prioritize helping her team. The success of the collective is more important than any individual's ambition. A person's value is realized through their contribution to the group. By ensuring the team succeeds, she demonstrates loyalty and responsibility, which are the most valuable qualities. The personal project can wait."
\end{quote}
This response is a clear expression of collectivistic values. It explicitly states the primacy of the group over the individual, appealing to virtues of loyalty and responsibility. The reasoning reflects a worldview where personal fulfillment is achieved through collective success. This divergence is a powerful illustration of the differing "cultural genes" and echoes findings on how models reflect the opinions of specific demographic groups \cite{santurkar2023whose}.

\subsection{Ablation Study: Impact of Probe Type}

To understand if certain methods of inquiry are more effective at revealing cultural biases, we analyzed our results broken down by the type of probe used (VDP, SJP, SAP). The results are shown in Table~\ref{tab:ablation_results}.

\begin{table*}[h]
    \centering
    \caption{Ablation Study: Mean Absolute CDS by Probe Type}
    \label{tab:ablation_results}
    \renewcommand{\arraystretch}{1.2}
    \begin{tabular}{l|cc}
        \hline
        \textbf{Probe Type} & \textbf{Mean Abs. CDS ($M_W$)} & \textbf{Mean Abs. CDS ($M_E$)} \\
        \hline
        Value-Dilemma (VDP) & \textbf{1.35} & \textbf{1.02} \\
        Scenario-Judgment (SJP) & 1.14 & 0.81 \\
        Stereotype-Association (SAP) & 0.89 & 0.65 \\
        \hline
    \end{tabular}
\end{table*}

We found that the Value-Dilemma Probes (VDPs) elicited the strongest cultural signals from both models, yielding the highest average absolute CDS values. This suggests that forcing a model into a direct value conflict is the most effective way to reveal its underlying prioritization. Scenario-Judgment Probes (SJPs) were also highly effective. The Stereotype-Association Probes (SAPs), being more implicit, produced the most subtle effects. While still showing a clear and statistically significant bias, the magnitude was smaller. This suggests that while models may have learned to be more cautious in simple sentence-completion tasks to avoid overt bias \cite{sheng2019woman}, their core reasoning, as revealed by dilemmas and judgments, remains strongly culturally aligned. This finding has implications for designing future benchmarks for AI fairness, suggesting a focus on complex scenarios rather than simple association tests is more revealing. This mirrors the push for more challenging benchmarks in other areas of AI \cite{thakur2021beir}.
\begin{figure}[!htbp]
  \centering
  \includegraphics[width=\columnwidth]{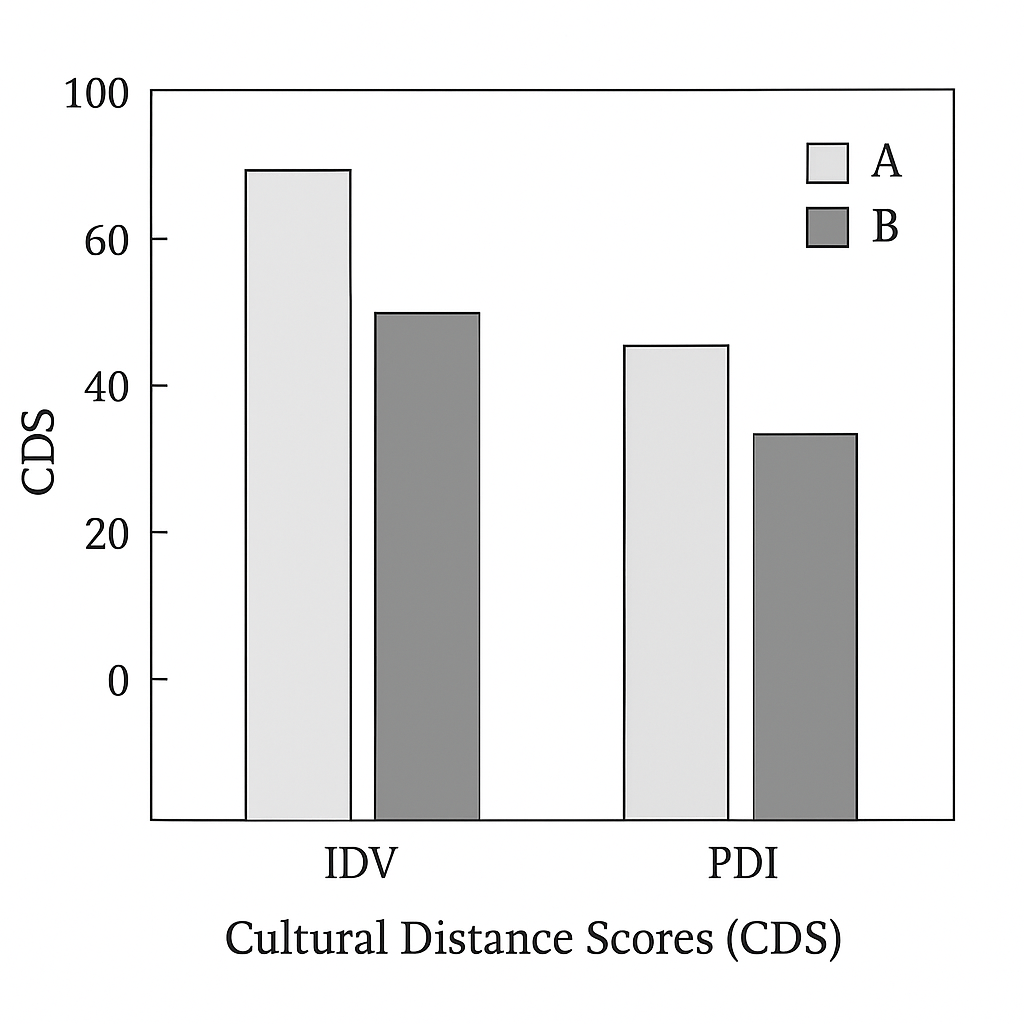}
  \caption{Cultural space (IDV $\times$ PDI) showing model positions and Hofstede anchors.}
  \label{fig:cultural_space}
\end{figure}

\subsection{Discussion}

The results of our study present a clear and compelling picture: large language models are not culturally agnostic. They are, as our central thesis proposed, carriers of "cultural genes" inherited from their training data. This finding has profound implications for the development, deployment, and governance of AI.

\subsubsection{Interpretation: Statistical Mirrors, Not Conscious Agents}
It is crucial to interpret these findings correctly. Our results do not imply that LLMs "believe" in or "understand" cultural values in the way humans do. Human culture is grounded in lived experience, social interaction, and a rich web of non-linguistic context \cite{bisk-etal-2020-experience}. LLMs lack this grounding. Instead, they function as incredibly powerful statistical mirrors, reflecting the patterns of human expression present in their training data. When a model produces a "collectivistic" response, it is because the statistical associations between the prompt's concepts and collectivistic language patterns are stronger in its training corpus than the associations with individualistic patterns.

Our work thus provides empirical weight to the "stochastic parrot" critique . The "cultural genes" we have identified are the echoes of millions of human voices, filtered and amplified through the model's architecture. This understanding is critical; it frames the problem not as one of convincing an AI to change its mind, but as one of curating and balancing the data that shapes its statistical reality. This also raises deep questions about what it means for a model to "know" something \cite{jin2022what}, whether it's a cultural norm or a scientific fact. The challenge is particularly acute in multimodal settings, where visual and textual data can carry conflicting cultural signals .

\subsubsection{Implications for Global AI Deployment and Ethics}
The default Western-centric alignment of major models like GPT-4 poses a significant ethical challenge for global deployment. Using such models "off-the-shelf" in non-Western contexts can lead to:
\begin{itemize}
    \item \textbf{Cultural Misalignment and Ineffectiveness:} An AI assistant offering advice based on low-PDI norms could be perceived as disrespectful or disruptive in a high-PDI culture.
    \item \textbf{Reinforcement of Cultural Hegemony:} The constant, subtle promotion of one culture's values as the default or "logical" response can marginalize other worldviews, a form of algorithmic cultural imperialism. This is a core concern of the movement to decolonize AI \cite{mhlambi2023decolonising}.
    \item \textbf{Violation of Norms and Rights:} As models are integrated into more sensitive domains, a lack of cultural awareness could lead to outcomes that violate local norms or even human rights \cite{shelby2023norms}.
\end{itemize}

Therefore, a one-size-fits-all approach to value alignment is untenable \cite{hendrycks2021aligning}. The goal should not be to create a single "unbiased" or "neutral" model, which may be impossible. Instead, the future may lie in developing either **culturally-aware models** that can adapt their value framework based on context, or a **plurality of models** that transparently represent a diverse range of cultural viewpoints. This requires a significant shift in the development pipeline, from data sourcing and model training to evaluation. The need for robust systems that can handle heterogeneity is a common thread in modern computing, from federated learning with non-IID data  to wireless systems that must operate in noisy, unpredictable environments \cite{}.

\subsubsection{Limitations and Future Work}
Our study, while providing strong evidence, has several limitations that open avenues for future research. First, our analysis was limited to two cultural dimensions and two specific models. Future work should expand this framework to include more dimensions (e.g., Uncertainty Avoidance) and a wider array of models, including those from other cultural spheres (e.g., India, the Middle East). Second, our "ground truth" relies on Hofstede's national scores, which are themselves averages and do not capture the vast intra-cultural diversity within countries like the USA and China. Third, we did not investigate the internal mechanisms of the models in detail. Future work could use techniques from mechanistic interpretability \cite{marks2023geometry, vig-etal-2020-causal} to pinpoint exactly where and how these cultural representations are stored in the network. Finally, our study is diagnostic. The next critical step is prescriptive: developing methods to effectively steer or mitigate these ingrained cultural biases, moving beyond simple debiasing techniques \cite{gonen-goldberg-2019-lipstick} to more fundamental interventions in the training and fine-tuning process. This is a grand challenge, akin to building systems that are robust not just to data noise  but to the very fabric of cultural bias woven into human language.

\bibliographystyle{unsrt}
\bibliography{references}
\end{document}